\documentclass[runningheads]{llncs}

\usepackage{accv}

\usepackage{accvabbrv}

\usepackage{graphicx}
\usepackage{booktabs}

\usepackage[accsupp]{axessibility}  % Improves PDF readability for those with disabilities.

\usepackage{pifont}
\usepackage{siunitx}
\usepackage{import}
\usepackage{colortbl}

\definecolor{mygreen}{RGB}{0,128,0}
\definecolor{lightgreen}{rgb}{0.7,1.0,0.7}
\definecolor{lightred}{rgb}{1.0,0.7,0.7}
\definecolor{accvblue}{rgb}{0.12,0.49,0.85}

\newcommand{\cmark}{\ding{51}}
\newcommand{\xmark}{\ding{55}}

\usepackage{hyperref}

\usepackage{orcidlink}

\begin{document}

\title{GRIP: Gaussian Rendering as a Cross-Modal Bridge for Image-to-Point Cloud Registration} 

\titlerunning{GRIP}
% % Include the authors' OCRID for the camera-ready version, if at all possible.
% Second Author\inst{2,3}\orcidlink{1111-2222-3333-4444} \and
% Third Author\inst{3}\orcidlink{2222--3333-4444-5555}}

\author{Karim Slimani\inst{1}\orcidID{0009-0007-4791-7173} \and
Catherine Achard\inst{1}\orcidID{0000-0002-5790-0830}
\and
Eric Marchand \inst{2}\orcidID{0000-0001-7096-5236} 
\and
Brahim Tamadazte\inst{1}\orcidID{0000-0002-4668-3092}}

\authorrunning{K.~Slimani et al.}

\institute{ISIR, Sorbonne Univ., CNRS , INSERM, Paris, France \\
Corresponding Author: \email{karim.slimani@isir.upmc.fr}
\and
Univ Rennes, Inria, CNRS, Irisa, Rennes, France
}

% % TODO FINAL: Replace with your institution list.
% Springer Heidelberg, Tiergartenstr.~17, 69121 Heidelberg, Germany

\maketitle

\begin{abstract}
This paper introduces \emph{GRIP}, a pose-conditioned refinement framework for pixel-to-point matching and 2D to 3D registration. Given an initial coarse pose estimate, \emph{GRIP} addresses the structural mismatch between grid based image descriptors and unordered point cloud descriptors by softly rendering learned 3D point features onto the image grid through Gaussian feature splatting. The rendered point derived feature map is then fused with image features by a pixel aligned transformer, enabling visual semantic and geometric cues to interact in a shared 2D representation. The refined features are decoded and propagated to finer resolutions for dense correspondence estimation and final pose refinement. Experiments on RGB D Scenes V2 and 7 Scenes demonstrate state of the art inlier ratio and competitive registration recall, with stronger performance under stricter evaluation thresholds.

\end{abstract}

\section{Introduction}\label{sec_intro}
% -----------
%
Image-to-point cloud registration is a fundamental task in computer graphics, computer vision, and robotics, with applications including robotic navigation~\cite{wang2025end}, 3D reconstruction~\cite{chen2025high}, and augmented reality~\cite{marchand2015pose,liu2020learning}. It aims to estimate the rigid transformation that registers a 3D point cloud with a partially overlapping 2D camera image.

This problem is highly challenging due to the data's heterogeneous nature. Captured through different modalities, images primarily provide visual cues such as color, texture, and semantics, whereas point clouds provide geometric cues such as spatial structure and local properties. Traditional 2D-3D registration methods typically follow a detect-then-match pipeline composed of three main steps: repeatable keypoints detection, pixel-to-point correspondence estimation, and transformation estimation. In these methods, keypoint detection and correspondence search rely on visual saliency and local geometric structures, while robust estimators, most notably PnP-RANSAC~\cite{lepetit2009ep,ransac1981}, are used to recover the final rigid pose. For a long time, such methods dominated cross-modal registration. However, challenging conditions such as sparse data, limited overlap, and large viewpoint changes make repeatable keypoint detection across these two different domains difficult, thereby degrading performance. To overcome these limitations, 2D3D-MATR~\cite{li20232d3d} introduced a detection-free pipeline. Inspired by state-of-the-art methods in image matching~\cite{sun2021loftr} and 3D point cloud registration~\cite{yu2021cofinet,geotransformer}, 2D3D-MATR divides the matching process into two complementary levels: coarse-level image patch and point-node matching and dense pixel-to-point matching. This paradigm inspired recent learning methods~\cite{li20232d3d,wu2024diff,mu2025diff2i2p,cheng2026rethinking} which adopted the same coarse-to-fine matching strategy coupled with PnP-RANSAC for scene registration~\cite{7scenes,rgbdv2}.

Despite these advances, detection-free 2D-3D matching still relies largely on direct feature comparison between two heterogeneous representations. In these methods, image descriptors are typically extracted using Feature Pyramid Network (FPN)~\cite{fpn} and ResNet~\cite{resnet} backbones, and therefore lie on a regular 2D grid that primarily encodes visual appearance and semantic context. In contrast, point cloud descriptors are predicted using backbones such as KPFCNN~\cite{thomas2019kpconv}, which encode geometric attributes in an unorganized 3D domain. Although existing methods~\cite{li20232d3d,wu2024diff,mu2025diff2i2p} adopt transformers to model cross-modal interactions at the coarsest level, the two modalities remain structurally misaligned during matching: image descriptors are arranged on a dense 2D grid, whereas point cloud descriptors are indexed over an unordered 3D points. As a result, these methods enhance contextual communication but do not explicitly unify the representation space in which cross-modal matching is performed. Moreover, since transformers are introduced only at the coarse level~\cite{li20232d3d,wu2024diff} and outside the backbone, fine-level features from the two modalities fail to capture cross-modal cues. This modality gap makes dense correspondence estimation challenging, especially under partial overlap, occlusions, and imperfect coarse alignment. Furthermore, once an initial pose is available, it provides a valuable geometric cue that can directly relate 3D points to image locations. However, existing coarse-to-fine pipelines do not fully exploit this pose to construct an image-aligned point cloud representation before refining the matching process.

In this paper, we present \emph{GRIP}, a pose conditioned refinement framework for image to point cloud registration based on feature splatting~\cite{kerbl20233d, wang2024pfgs}. Starting from an initial coarse pose estimate, \emph{GRIP} renders learned 3D point features into the image plane, producing an image aligned representation of the point cloud. This representation serves as a cross modal bridge, enabling pixel aligned interaction between visual features and 3D derived geometric features prior to final correspondence estimation. \emph{GRIP} therefore changes the interaction space before final matching: instead of refining descriptors only in their original heterogeneous domains, it first renders learned 3D features onto the image plane and then performs cross modal interaction within a shared pixel aligned representation. This representation is subsequently used to refine dense correspondences and recover the final pose, as detailed in Section~\ref{sec_method}.

In a nutshell, the main contributions of this paper are as follows:
% \textcolor{blue}{A peaufiner}
\begin{itemize}
    \item We introduce a Point-to-Pixel feature rendering module that converts unordered point-cloud descriptors into an image-aligned feature map.
    % \item We identify that recent Gaussian splatting rendering enables a continuous representation of point clouds.
    % \item We introduce a Point-to-Pixel Feature Renderer that exploits Gaussian Splatting to project coarse point-cloud features onto a coarse image grid, transforming an unordered point-cloud representation into a structured feature map aligned with the image domain.
    \item We propose a Pixel-Aligned Interaction Transformer that benefits from this intermediate representation and bidirectionally fuses geometric and visual-semantic cues on a shared 2D grid.
    \item We design a hierarchical propagation strategy that injects the fused cross-modal representation into both image and point-cloud decoders, improving coarse and dense 2D-3D matching and pose refinement.
    \item We achieve state-of-the-art matching inlier ratio and competitive registration recall, with stronger performance under stricter thresholds.
\end{itemize}

%
% ---------------
\section{Related works}
% %
% % \TODO{pour gain de place, réduire au minimum le paragraphe suivant }\\
% Recent 2D-3D registration methods draw inspiration from both image matching and 3D point cloud registration. Before deep learning, image matching mainly relied on handcrafted local features~\cite{rublee2011orb,sift}, which often struggled in textureless regions and large-disparity settings due to their limited local support. Early learning-based methods~\cite{detone2018superpoint,dusmanu2019d2} improved descriptor discriminativeness with CNNs, while SuperGlue~\cite{sarlin20superglue} introduced transformer-based feature matching~\cite{vaswani2017attention} to aggregate global context for more reliable correspondences.
% Similarly, traditional point cloud registration methods relied on hand-crafted geometric features, with Iterative Closest Point (ICP)~\cite{besl1992method} and its derivatives~\cite{goicp,fasticp} serving as a classical reference. With deep learning, DCP~\cite{dcp} integrated learned point features using a DGCNN backbone~\cite{dgcnn}, a transformer module~\cite{vaswani2017attention}, and a weighted SVD~\cite{soft_svd} for pose estimation. More recently, coarse-to-fine schemes have become a common paradigm~\cite{sun2021loftr,yu2021cofinet,geotransformer}, applying transformer-based interactions at coarse resolution and then propagating the matched information to finer levels.

% \subsection{Image and Point Cloud Registration}

Recent 2D-3D registration methods draw inspiration from both image matching and 3D point cloud registration. Image matching evolved from handcrafted descriptors~\cite{rublee2011orb,sift} to learning-based methods~\cite{detone2018superpoint,dusmanu2019d2}, with transformer-based architectures such as SuperGlue~\cite{sarlin20superglue} improving correspondence estimation through global context aggregation. Similarly, point cloud registration progressed from ICP and its derivatives~\cite{besl1992method,goicp,fasticp} to deep learning approaches such as DCP \cite{dcp}, before recent coarse-to-fine transformer frameworks became the dominant paradigm \cite{sun2021loftr,yu2021cofinet,geotransformer}.

Building upon these advances, 2D-3D registration has evolved from detect-and-match pipelines based on handcrafted features~\cite{li2012worldwide} to recent detection-free learning frameworks~\cite{li20232d3d,wu2024diff}. Inspired by learning-free predecessors~\cite{sattler2016efficient}, which extract local features using SIFT~\cite{sift}, 2D3DMatchNet~\cite{feng20192d3d} introduces a learning-based approach relying on PointNet~\cite{qi2016pointnet} and CNNs to jointly predict descriptors for 2D and 3D keypoints. Moving beyond handcrafted detectors, P2-Net~\cite{wang2021p2} proposes a dual-fully-convolutional framework that maps 2D and 3D inputs into a shared latent space to jointly detect and describe keypoints. However, as highlighted by 2D3D-MATR~\cite{li20232d3d}, existing inter-modality methods either rely on scene-specific coordinate regression, thereby limiting generalization to novel scenes, or follow detect-then-match pipelines that are hindered by unstable cross-modal keypoint repeatability and low inlier ratios. 

2D3D-MATR~\cite{li20232d3d} introduced a detection-free coarse-to-fine framework for 2D-3D registration. Multi-resolution image and point cloud features interact through a transformer~\cite{bello2019attention} at the coarsest level to establish correspondences that subsequently guide dense pixel-to-point matching. The good performance of 2D3D-MATR on challenging indoor benchmarks such as RGB-D Scenes V2~\cite{rgbdv2} and 7-Scenes~\cite{7scenes} inspired several subsequent methods to adopt a similar coarse-to-fine design. In particular, Diff-Reg~\cite{wu2024diff} extends 2D3D-MATR with a diffusion-based matching framework that preserves the original pipeline and employs a transformer-based denoising module to iteratively refine the coarse matching matrix. Diff$^2$I2P~\cite{mu2025diff2i2p} further leverages a depth-conditioned diffusion model to distill cross-modal knowledge and bridge the gap between 2D and 3D features while maintaining the same coarse-to-fine paradigm. More recently, R$^{23}$Net~\cite{cheng2026rethinking} addressed the issue of non-overlapping and low-quality matching regions inherent to coarse-to-fine 2D-3D registration by first identifying informative regions in both the image and the point cloud through a reinforcement-learning-based High-Value Zone Reinforced Selection module. 

In contrast to these approaches, which mainly refine matching scores or cross-modal descriptors within the original feature domains, \emph{GRIP} explicitly modifies the representation space by rendering 3D descriptors onto the image plane prior to cross-modal interaction.
%%%%%%%%%%%%%%%%%
%
% -------------------
\section{Method}\label{sec_method}

%
% -------------------
\subsection{Problem Statement}
% -------------------
%
Let $\mathbf{I}$ be an RGB image and $\mathbf{P} = \{\mathbf{p}_1, \dots, \mathbf{p}_N\} \subset \mathbb{R}^{3}$ be a point cloud. 
We assume that $\mathbf{I}$ and $\mathbf{P}$ are partially overlapping, such that a set of valid 2D-3D correspondences exists:
$\mathcal{C}^{gt}=\{(\mathbf{p}_k,\mathbf{u}_k)\mid \mathbf{p}_k\in\mathbb{R}^{3},\mathbf{u}_k\in\mathbb{R}^{2}\}$ where $\mathbf{u}_k$ are the 2D pixel coordinates. 
The goal of 2D-3D registration is to estimate the rigid transformation $\mathbf{\hat{T}} (\mathbf{\hat{R}},\mathbf{\hat{t}}) \in \mathrm{SE(3)}$, 
that minimizes the reprojection error between these correspondences:

\begin{equation}
\mathbf{\hat{R}},\mathbf{\hat{t}} = \underset{\mathbf{R}, \mathbf{t}}{\arg\min}
\sum_{(\mathbf{p}_k,\mathbf{u}_k)\in\mathcal{C}^{gt}}
\big\| 
\mathbf{u}_k - \pi\big(\mathbf{R}\mathbf{p}_k + \mathbf{t};\mathbf{K}\big) 
\big\|_2^2
\label{eq:2d3d_reg}
\end{equation}
where $\pi(~\cdot~;\mathbf{K}):\mathbb{R}^{3}\rightarrow\mathbb{R}^{2}$ denotes the perspective projection from 3D camera coordinates to 2D image coordinates using the camera intrinsic matrix $\mathbf{K}$.
% -------------------
%

The proposed method follows a two-stage strategy. Stage~1 estimates an initial transformation~\ref{subsec_stage1}, while the main contribution lies in Stage~2, which performs pose-conditioned rendering-based refinement. Stage~2 comprises \textbf{Point-to-Pixel Feature Rendering}, \textbf{Pixel-Aligned Interaction}, \textbf{Hierarchical Cross-Modal Feature Propagation}, and final correspondence and pose estimation, as detailed in~\ref{subseq_fet_rend}.

% The proposed method follows a two-stage strategy. The first stage serves as an initialization module, estimating an initial transformation between the image and the point cloud. The main contributions of \emph{GRIP}, as mentioned in the introduction, lie in the second stage, which performs rendering-based refinement. Given the initial transformation, we introduce a \textbf{Point-to-Pixel Feature Renderer} that softly splats coarse point cloud features onto the coarse image grid. The resulting rendered point-derived feature map and the image feature map are then processed by a \textbf{Pixel-Aligned Interaction Transformer}, which bidirectionally fuses geometric and visual-semantic cues on a shared 2D grid. The fused representations are propagated to finer resolutions through a \textbf{Hierarchical Cross-Modal Feature Propagation} block. Finally, refined coarse and dense correspondences are estimated, and the \textbf{final correspondences and pose estimation} are recovered using either score-guided PnP-RANSAC or a weighted PnP solver~\cite{chen2022epro}.
%
\begin{figure*}
\centerline{\includegraphics[width=1\columnwidth]{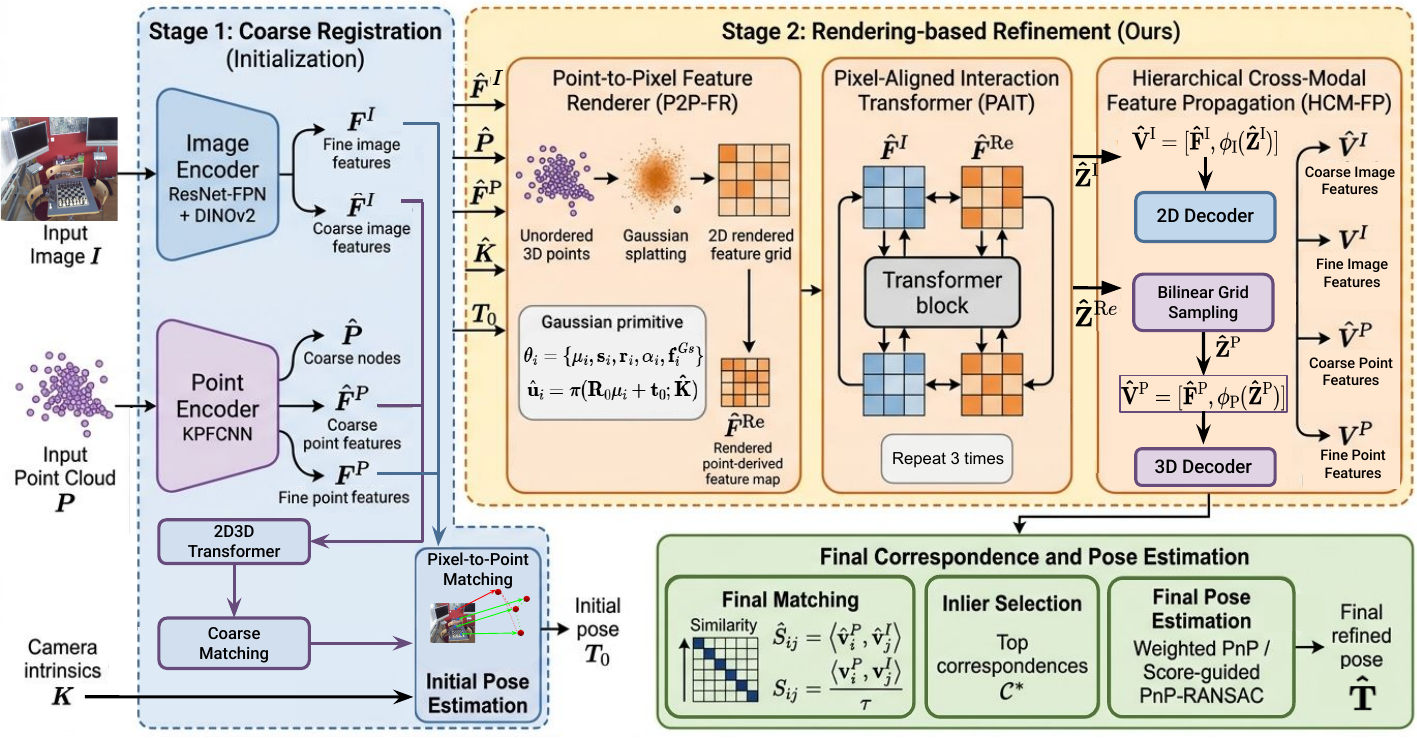}}
\caption{Summary of the proposed \emph{GRIP} method.}
\label{fig_pipeline}
\end{figure*}
%
% -----------
\subsection{Coarse Registration (Initialization)}\label{subsec_stage1}

Given an image $\mathbf{I}\in\mathbb{R}^{H\times W\times 3}$ and a point cloud $\mathbf{P}\in\mathbb{R}^{N\times 3}$, the first stage estimates an initial transformation between the two modalities. It follows the coarse-to-fine matching paradigm of 2D3D-MATR~\cite{li20232d3d} and provides the initialization for the proposed rendering-based refinement module. Multi-resolution features are extracted using modality-specific backbones: KPFCNN~\cite{thomas2019kpconv} for the point cloud, and a ResNet-FPN backbone~\cite{resnet,fpn} for the image, with the deepest image features augmented by DINOv2 descriptors~\cite{oquab2023dinov2}.

At the coarse level, we denote the image-patch and point-node features by $\hat{\mathbf{F}}^{\mathrm I}\in\mathbb{R}^{\hat{H}\times\hat{W}\times\hat{C}}$ and $\hat{\mathbf{F}}^{\mathrm P}\in\mathbb{R}^{\hat{N}\times\hat{C}}$, respectively. At the fine level, the corresponding features are denoted by $\mathbf{F}^{\mathrm I}\in\mathbb{R}^{H\times W\times C}$ and $\mathbf{F}^{\mathrm P}\in\mathbb{R}^{N\times C}$. 
% Here, $H$ and $W$ are the image dimensions, $N$ is the number of input points, $\hat{H}\times\hat{W}$ and $\hat{N}$ define the coarse image and point-cloud resolutions, and $C$ and $\hat{C}$ are the fine- and coarse-level feature dimensions.
%
A cross-modal transformer~\cite{li20232d3d} refines the coarse features $\hat{\mathbf{F}}^{\mathrm I}$ and $\hat{\mathbf{F}}^{\mathrm P}$ to produce context-aware descriptors $\hat{\mathbf{H}}^{\mathrm I}\in\mathbb{R}^{\hat{H}\hat{W}\times\hat{C}}$ and $\hat{\mathbf{H}}^{\mathrm P}\in\mathbb{R}^{\hat{N}\times\hat{C}}$. Coarse image-patch-to-point-node correspondences are obtained by computing the pairwise similarity:
\begin{equation}
\hat{\mathbf{S}}^{c}_{ij} =
\left\langle
\hat{\mathbf{h}}^{\mathrm P}_{i},
\hat{\mathbf{h}}^{\mathrm I}_{j}
\right\rangle ,
\end{equation}
where $\hat{\mathbf{h}}^{\mathrm P}_{i}$ denotes the $i$-th coarse point feature and $\hat{\mathbf{h}}^{\mathrm I}_{j}$ denotes the $j$-th image patch feature. A top-$k$ selection is then applied to $\hat{\mathbf{S}}^{c}$, yielding a set of coarse correspondences $\hat{\mathcal{C}}=
\left\{(\hat{\mathbf{p}}_{m},\hat{\mathbf{u}}_{m})\right\}_{m=1}^{N_c}$, where each coarse pair links an image patch $\hat{\mathbf{u}}_{m}$ to a point node $\hat{\mathbf{p}}_{m}$, and $N_c$ is the number of selected correspondences.

The coarse correspondences are then used to restrict dense pixel-to-point matching to local patch pairs. For each coarse pair $(\hat{\mathbf{p}}_{m},\hat{\mathbf{u}}_{m})$, the associated local 3D points and local image pixels are retrieved from the fine-resolution representations. Dense similarities are computed between their fine-level descriptors, and a mutual top-$k$ selection is applied to obtain a set of dense correspondences $\mathcal{C}_{m}$. The final dense correspondence set is obtained by aggregating all local matches as $\mathcal{C} = \bigcup_{m=1}^{N_c} \mathcal{C}_{m}$. 

In our experiments, the initial transformation $\mathbf{T}_{0}\in \mathrm{SE}(3)$ is then estimated from $\mathcal{C}$ using PnP-RANSAC~\cite{lepetit2009ep,ransac1981} with a maximum of $2{,}000$ iterations. This transformation serves as the geometric guide for the feature rendering-based refinement stage described below.
%
% ------------
\subsection{Feature Rendering-based Refinement}~\label{subseq_fet_rend}
A key design principle of the proposed \emph{GRIP} is that reliable 2D-3D correspondences should be refined via an intermediate representation that spatially aligns image and point cloud features, rather than only through direct cross-modal descriptor comparison. 
Accordingly, \emph{GRIP} consists of four main components: i) the \textbf{Point-to-pixel Feature Renderer} converts unordered point-cloud descriptors into an image-aligned feature map via pose-aware feature splatting using the initial pose $\mathbf{T}_{0}$. ii) The \textbf{Pixel-Aligned Interaction Transformer} refines image features and rendered 3D-derived features through bidirectional cross-modal attention on a shared 2D grid. iii) The \textbf{Hierarchical Cross-Modal Feature Propagation} maps the refined rendered features back to the sparse 3D nodes, yielding updated point cloud descriptors enriched with visual information. iv)  \textbf{Matching and Pose Estimation} uses the refined descriptors to recover improved dense 2D-3D correspondences and estimate the final rigid transformation.
\subsubsection{Point-to-pixel feature renderer}
Given coarse point features $\hat{\mathbf{F}}^{\mathrm P}\in\mathbb{R}^{\hat{N}\times\hat{C}}$ associated with points $\hat{\mathbf{P}}\in\mathbb{R}^{\hat{N}\times3}$, the Point-to-Pixel Feature Renderer uses the initial pose $\mathbf{T}_0$ to softly splat point features onto the coarse image grid, transforming an unordered point representation into an image-aligned feature map~\cite{kerbl20233d,wang2024pfgs}. This soft and differentiable assignment is particularly beneficial under imperfect pose initialization. In contrast to hard projection, which assigns each 3D feature to a single pixel and is therefore prone to misalignment under small pose perturbations, Gaussian splatting distributes feature responses over a local image neighbourhood, enabling nearby tokens to retain informative geometric cues. For each coarse 3D point $\hat{\mathbf{p}}_{i}\in\hat{\mathbf{P}}$, we instantiate one renderable Gaussian primitive, containing the attributes required by the splatting renderer: a 3D Gaussian shape, an opacity value, and a feature vector to be projected onto the image grid. The spatial support of this primitive is an anisotropic 3D Gaussian $\mathcal{G}_{i}$ defined by a center $\mathbf{\mu}_{i}$ and a covariance matrix $\mathbf{\Sigma}_{i}$. Following anisotropic Gaussian splatting~\cite{kerbl20233d,wang2024pfgs}, the covariance matrix is parameterized by a rotation matrix $\mathbf{Q}_{i}$ and a scaling matrix $\mathbf{\Lambda}_i$ as $\mathbf{\Sigma}_{i}=\mathbf{Q}_{i}\mathbf{\Lambda}_i\mathbf{\Lambda}_i^{\top}\mathbf{Q}_{i}^{\top}$.

The primitive parameters can then be defined as $\mathbf{\theta}_{i}=\{\mathbf{\mu}_{i},\mathbf{s}_i,\mathbf{r}_{i},\alpha_{i},\mathbf{f}^{Gs}_{i}\}$.
Here, $\mathbf{\mu}_{i}=\hat{\mathbf{p}}_{i}+\Delta\mathbf{p}_{i}$ is the Gaussian center, where $\Delta\mathbf{p}_{i}$ is a learned offset. 
The vector $\mathbf{s}_i$ defines the anisotropic scale, with $\mathbf{\Lambda}_i=\operatorname{diag}(\mathbf{s}_i)$, while $\mathbf{r}_{i}$ parameterizes the Gaussian rotation matrix $\mathbf{Q}_{i}$. 
The scalar $\alpha_{i}$ denotes the opacity, and $\mathbf{f}^{Gs}_{i}\in\mathbb{R}^{C_{Gs}}$ is the feature vector carried by the primitive and splatted onto the image grid.  All these attributes are predicted from the input point feature $\hat{\mathbf{f}}^{\mathrm P}_{i}$ as:

\begin{equation}
\mathbf{f}^{Gs}_{i}=\hat{\mathbf{f}}^{\mathrm P}_{i}\mathbf{W}
\end{equation}
where $\mathbf{W}\in\mathbb{R}^{\hat{C}\times{C}_{Gs}}$ is a learned projection. The remaining primitive parameters are predicted as functions of the point descriptor $\hat{\mathbf{f}}^{\mathrm P}_{i}$:
{\small
\begin{equation}
\Delta\mathbf{p}_{i}=h_\theta^{p}(\hat{\mathbf{f}}^{\mathrm P}_{i}),\quad
\mathbf{r}_{i}=h_\theta^{r}(\hat{\mathbf{f}}^{\mathrm P}_{i}),\quad
\mathbf{s}_i=\operatorname{softplus}(h_\theta^{s}(\hat{\mathbf{f}}^{\mathrm P}_{i})),\quad
\alpha_{i}=\sigma(h_\theta^{\alpha}(\hat{\mathbf{f}}^{\mathrm P}_{i})).
\end{equation}
}
where $h_\theta^{p}:\mathbb{R}^{\hat{C}}\rightarrow\mathbb{R}^{3}$ predicts the center offset, $h_\theta^{r}:\mathbb{R}^{\hat{C}}\rightarrow\mathbb{R}^{6}$ predicts a continuous 6D rotation representation following~\cite{zhou2019continuity}, $h_\theta^{s}:\mathbb{R}^{\hat{C}}\rightarrow\mathbb{R}^{3}$ predicts the anisotropic scale, and $h_\theta^{\alpha}:\mathbb{R}^{\hat{C}}\rightarrow\mathbb{R}$ predicts the opacity, with $\sigma(\cdot)$ denoting the sigmoid function. The offset prediction head $h_\theta^{p}$ is initialized to zero, and no explicit regularization is imposed on the predicted offsets. The predicted rotation parameter $\mathbf{r}_{i}$ is converted into a valid rotation matrix $\mathbf{Q}_{i}$, while $\mathbf{s}_i$ defines the diagonal scaling matrix $\mathbf{\Lambda}_i=\operatorname{diag}(\mathbf{s}_i)$.

At this point, one can exploit the known camera intrinsics $\mathbf{K}$ and the initial pose $\mathbf{T}_{0}$ to build an organized representation of the point cloud $\hat{\mathbf{F}}^{Re}\in\mathbb{R}^{\hat{H}\times\hat{W}\times{C}_{Gs}}$ by rendering the attributes of each primitive into the coarse image plane:
\begin{equation}
\hat{\mathbf{F}}^{Re} = 
\mathbf{Re}\left(
\{\mathbf{\mu}_{i},\mathbf{Q}_{i},\mathbf{\Lambda}_i,\alpha_{i},\mathbf{f}^{Gs}_{i}\}_{i=1}^{\hat{N}}\mid
\mathbf{\hat{K}},\mathbf{T}_{0}
\right).
\end{equation}
where $\mathbf{Re}$ denotes the Gaussian feature renderer~\cite{kerbl20233d,wang2024pfgs}. Before rendering, points with invalid camera-frame depth are discarded, and primitives projected far outside the image domain are filtered out. The renderer projects each remaining primitive into the image plane using the scaled coarse intrinsics $\mathbf{\hat{K}}$, derived from $\mathbf{K}$, and the initial pose $\mathbf{T}_{0}$, computes its projected Gaussian footprint, and alpha-composites the feature vectors carried by the visible primitives, as in~\cite{wang2024pfgs}.
The effective opacity contribution of primitive $\mathbf{\theta}_{i}$ at location $\hat{\mathbf{u}}$ is defined as:
\begin{equation}
g_{i}(\hat{\mathbf{u}})
=
\alpha_{i}
\exp\left(
-\frac{1}{2}
(\hat{\mathbf{u}}-\hat{\mathbf{u}}_{i})^{\top}
(\mathbf{\Sigma}^{2D}_{i})^{-1}
(\hat{\mathbf{u}}-\hat{\mathbf{u}}_{i})
\right)
\end{equation}
where $\hat{\mathbf{u}}_{i}=\pi(\mathbf{R}_{0}\mathbf{\mu}_{i}+\mathbf{t}_{0};\mathbf{\hat{K}})$ is the projected center of the $i$-th primitive with the rotation  $\mathbf{R}_{0}$  and the translation $\mathbf{t}_{0}$ from $\mathbf{T}_{0}$ and 
$\mathbf{\Sigma}^{2D}_{i}$ denotes the projected 2D covariance of the $i$-th Gaussian primitive. The rendered feature is then computed as:
\begin{equation}
\hat{\mathbf{F}}^{Re}(\hat{\mathbf{u}})
=
\sum_{i\in\mathcal{N}(\hat{\mathbf{u}})}
T_{i}(\hat{\mathbf{u}})
g_{i}(\hat{\mathbf{u}})
\mathbf{f}^{Gs}_{i}
\end{equation}
where $\mathcal{N}(\hat{\mathbf{u}})$ denotes the set of projected primitives contributing to location $\hat{\mathbf{u}}$ and $T_{i}$ the accumulated transmittance is defined as:
\begin{equation}
T_{i}(\hat{\mathbf{u}})
=
\prod_{j<i}
\left(1-g_{j}(\hat{\mathbf{u}})\right)
\end{equation}

The primitives are sorted in front-to-back order according to their depth in the camera view.  The resulting feature map $\hat{\mathbf{F}}^{Re}$ is thus an image-aligned representation of the point cloud features. It can therefore interact with the image backbone features $\hat{\mathbf{F}}^{\mathrm I}$. The shared coarse grid enables pixel-aligned cross-attention between rendered point features and 2D image features, as detailed below.
\subsubsection{Pixel-aligned interaction transformer}
The main purpose of this central block is to exploit the organized structure of the rendered point cloud feature map $\hat{\mathbf{F}}^{Re}\in\mathbb{R}^{\hat{H}\times\hat{W}\times{C}_{Gs}}$ to enable information exchange between the two modalities. The rendered descriptor $\hat{\mathbf{F}}^{Re}$ captures local geometric structure in the point cloud while the image feature map $\hat{\mathbf{F}}^{\mathrm I}\in\mathbb{R}^{\hat{H}\times\hat{W}\times\hat{C}}$  mainly encodes visual appearance and semantic context. Since both representations are organized on the same coarse image grid, they can interact in a shared pixel-aligned space. This interaction allows the rendered point cloud features to integrate visual context from the image, while the image features, in turn, are enhanced with geometry-aware cues from the point cloud. 

To this end, we map both feature maps into a shared feature dimension $\hat{C}$. The rendered point cloud feature map and the image feature map are then flattened into token sequences and passed through linear projections as:
\begin{equation}
{}^{1}\hat{\mathbf{F}}^{Re}=\operatorname{Flatten}(\hat{\mathbf{F}}^{Re})\mathbf{W}^{\mathrm P},\quad
{}^{1}\hat{\mathbf{F}}^{\mathrm I}=\operatorname{Flatten}(\hat{\mathbf{F}}^{\mathrm I})\mathbf{W}^{\mathrm I}
\end{equation}
where $\mathbf{W}^{\mathrm P}\in\mathbb{R}^{\hat{C}_{Gs}\times \hat{C}}$ and 
$\mathbf{W}^{\mathrm I}\in\mathbb{R}^{\hat{C}_{I}\times \hat{C}}$ are learned modality-specific linear projection matrices, and ${}^{1}\hat{\mathbf{F}}^{Re},{}^{1}\hat{\mathbf{F}}^{\mathrm I}\in\mathbb{R}^{\hat{H}\hat{W}\times \hat{C}}$. Since both token sequences are defined on the same coarse image grid, we associate each token with its normalized 2D coordinate ${px}_{i}\in[-1,1]^{2}$, making the attention aware of the pixel location of each token while preserving their alignment on the same grid. These coordinates are encoded with a Fourier positional embedding~\cite{qin2023deep,mildenhall2021nerf}, denoted as $\gamma(\cdot)$, and projected to the same feature dimension:
\begin{equation}
\mathbf{e}_{i}=\mathbf{W}^{pos}\gamma({px}_{i})
\end{equation}

We perform bidirectional shifted cross-attention~\cite{liu2021swin}, \textit{i.e.}, from point cloud to image and from image to point cloud, to update both feature maps with the same algorithm. For instance, consider the point cloud-to-image direction, where image tokens query the rendered point cloud tokens. Let the output of one attention layer be the feature matrix $\mathbf{Z}\in\mathbb{R}^{\hat{H}\hat{W}\times \hat{C}}$. It is then given by:
\begin{equation}
\mathbf{Z}_{i}=
\sum_{j\in\mathcal{W}(i)}
a_{i,j}
\left(
{}^{1}\hat{\mathbf{f}}^{Re}_{j}\mathbf{W}^{V}
\right),
\end{equation}
where ${}^{1}\hat{\mathbf{f}}^{Re}_{j}\in\mathbb{R}^{\hat{C}}$ denotes the $j$-th rendered point-cloud token, \textit{i.e.}, the $j$-th row of ${}^{1}\hat{\mathbf{F}}^{Re}$, and $\mathcal{W}(i)$ denotes the local shifted window associated with token $i$, $\mathbf{W}^{V}$ is the value projection matrix, and $a_{i,j}$ denotes the attention weight between the $i$-th image token and the $j$-th rendered point cloud token within this window. The attention weights are computed as:
\begin{equation}\label{eq_attention}
a_{i,j}=
\operatorname{softmax}_{j}
\left(
\frac{
\left(
({}^{1}\hat{\mathbf{f}}^{\mathrm I}_{i}+\mathbf{e}_{i})\mathbf{W}^{Q}
\right)
\left(
({}^{1}\hat{\mathbf{f}}^{Re}_{j}+\mathbf{e}_{j})\mathbf{W}^{K}
\right)^{\top}
}
{\sqrt{C}}
\right)
\end{equation}
where $\mathbf{W}^{Q}$ and $\mathbf{W}^{K}$ are the query and key projection matrices, respectively. %This operation updates each image token by aggregating information from the rendered point cloud feature map. The reverse direction, which updates the rendered point cloud tokens using image features, is obtained symmetrically by swapping the roles of the two modalities. 
This bidirectional interaction enables image and rendered point-derived tokens to selectively incorporate both geometric and semantic information, helping the model capture repeated patterns across modalities and facilitating feature matching. The transformer outputs refined rendered point cloud and image feature maps, denoted as $\mathbf{\hat{Z}}^{Re}\in\mathbb{R}^{\hat{H}\times\hat{W}\times\hat{C}}$ and $\mathbf{\hat{Z}}^{\mathrm I}\in\mathbb{R}^{\hat{H}\times\hat{W}\times\hat{C}}$, respectively. These features are then decoded and propagated to finer resolutions, as detailed below.
\subsubsection{Hierarchical cross-modal feature propagation}
At this stage, we progressively propagate coarse point clouds and image feature maps to finer resolutions. 
On the image side, $\hat{\mathbf{Z}}^{\mathrm I}$ is first projected by a fusion Feed-Forward Network (FFN) and concatenated with the original coarse image descriptor $\hat{\mathbf{F}}^{\mathrm I}$ to produce $\mathbf{\hat{V}}^{\mathrm I} \in \mathbb{R}^{\hat{H} \hat{W}\times 2\hat{C}}$:
\begin{equation}
    \mathbf{\hat{V}}^{\mathrm I}
    =
    [\mathbf{\hat{F}}^{\mathrm I}, \phi_{\mathrm I}(\mathbf{\hat{Z}}^{\mathrm I})]
\end{equation}
where $[\cdot,\cdot]$ denotes channel-wise concatenation. The fusion function $\phi_{\mathrm I}$ is implemented as an FFN, consisting of a LayerNorm layer followed by two linear projections with a GELU non-linearity in between.  The resulting image features $\mathbf{\hat{V}}^{\mathrm I}$ are then fed into an FPN-style 2D decoder,  which progressively propagates enhanced information to finer image resolutions following the first-stage architecture described in Section~\ref{subsec_stage1}.

In the same spirit, we fuse the coarse point cloud descriptor $\hat{\mathbf{F}}^{\mathrm P}\in\mathbb{R}^{\hat{N}\times\hat{C}}$ with the Pixel-Aligned Interaction Transformer output $\hat{\mathbf{Z}}^{Re}\in\mathbb{R}^{\hat{H}\times\hat{W}\times\hat{C}}$. 
To do so, we first need to recover the feature vector associated with each coarse node $\hat{\mathbf{p}}_{i}$. 
Each node is projected onto the coarse image grid using the intrinsics and the initial transformation $\mathbf{T}_{0}$. 
Thereby, the corresponding feature vector for the node $\hat{\mathbf{p}}_{i}$ is aggregated from $\hat{\mathbf{Z}}^{Re}$ via bilinear grid sampling, which extracts features from the surrounding pixels of its 2D projection. 
The resulting point-wise feature map is denoted as $\hat{\mathbf{Z}}^{\mathrm P}\in\mathbb{R}^{\hat{N}\times\hat{C}}$. 
We then compute the fused point cloud features $\mathbf{\hat{V}}^{\mathrm P} \in \mathbb{R}^{\hat{N}\times 2\hat{C}}$ as:
\begin{equation}
    \mathbf{\hat{V}}^{\mathrm P}
    =
    [\mathbf{\hat{F}}^{\mathrm P}, \phi_{\mathrm P}(\mathbf{\hat{Z}}^{\mathrm P})]
\end{equation}
where the fusion function $\phi_{\mathrm P}$ shares the same architecture as $\phi_{\mathrm I}$.  The resulting coarse feature map $\mathbf{\hat{V}}^{\mathrm P}$ is then fed into a KPFCNN~\cite{thomas2019kpconv} decoder, following the first-stage architecture described in Section~\ref{subsec_stage1}. The dense image and point cloud feature maps, denoted as $\mathbf{V}^{\mathrm I}\in\mathbb{R}^{H\times W\times C}$ and $\mathbf{V}^{\mathrm P}\in\mathbb{R}^{N\times C}$, respectively, are used to guide the matching and registration processes, as explained below.
\subsubsection{Matching and pose estimation}
Given the refined coarse features $\hat{\mathbf{V}}^{\mathrm I}\in\mathbb{R}^{\hat{H} \hat{W}\times 2\hat{C}}$ and $\hat{\mathbf{V}}^{\mathrm P}\in\mathbb{R}^{\hat{N}\times 2\hat{C}}$, together with the decoded dense features $\mathbf{V}^{\mathrm I}\in\mathbb{R}^{H W\times C}$ and $\mathbf{V}^{\mathrm P}\in\mathbb{R}^{N\times C}$, we perform the same coarse to fine matching strategy as described in Section~\ref{subsec_stage1}. We first compute a coarse similarity matrix between the refined point cloud nodes and image patches: 
\begin{equation}
\hat{\mathbf{S}}_{ij}
=
\left\langle
\hat{{v}}^{\mathrm P}_{i},
\hat{{v}}^{\mathrm I}_{j}
\right\rangle
\end{equation}
where $\hat{{v}}^{\mathrm P}_{i}$ and $\hat{{v}}^{\mathrm I}_{j}$ denote the refined coarse descriptors of the $i$-th point cloud node and the $j$-th image patch, respectively. A masked top-$k$ selection is then applied to $\hat{\mathbf{S}}$ to obtain a set of coarse patch correspondences $\hat{\mathcal{C}}={(\hat{\mathbf{p}}_{m},\hat{\mathbf{u}}_{m})}_{m=1}^{N_{c}}$.

For each selected coarse correspondence, we retrieve the associated local image pixel and point cloud, and compute dense similarities between their decoded fine-level descriptors. The dense similarity between a point descriptor ${v}^{\mathrm P}_{i}$ and an image descriptor ${v}^{\mathrm I}_{j}$ is defined as
\begin{equation}
\mathbf{S}_{ij}
=
\frac{
\left\langle
{v}^{\mathrm P}_{i},
{v}^{\mathrm I}_{j}
\right\rangle
}{\tau}
\end{equation}
where $\tau$ is a temperature parameter. Mutual top-$k$ selection is then applied within each matched patch pair to recover the dense pixel-to-point correspondence set $\mathcal{C}^*$. Finally, the $2048$ highest-scoring correspondences are used for pose estimation. We evaluate two final pose estimators. The first is a score-guided PnP-RANSAC variant. Instead of sampling minimal sets uniformly~\cite{ransac1981}, each correspondence is assigned a confidence score derived from the dense similarity matrix $\mathbf{S}$. Correspondences are sorted by decreasing confidence, and minimal PnP sets are first sampled from high-confidence subsets. This prioritization enables reliable matches to be tested earlier, reducing the maximum number of RANSAC iterations to $500$, compared with the $50{,}000$ iterations used by the methods reported in Section~\ref{sec_expe}, such as~\cite{li20232d3d,huang2021predator,wu2024diff}. Each pose hypothesis is evaluated over all correspondences using the reprojection error with an $8$ pixel inlier threshold. The second variant is the weighted PnP solver of~\cite{chen2022epro}, which directly uses the correspondence confidence scores $\mathbf{S}$ to estimate the transformation in a small number of iterations, \textit{i.e.,} $3$ in our case.

\subsection{Implementation details}
The method is implemented in PyTorch and trained on an NVIDIA Tesla V100 GPU with 32GB of memory. For both stage 1 and 2 training, we adopt the same coarse-to-fine circle loss supervision commonly used in recent image-to-point cloud registration methods~\cite{li20232d3d,wu2024diff}. The network is trained with the Adam optimizer~\cite{kingma2014adam} using a learning rate of $10^{-4}$, which is decayed by a factor of $0.95$ at each epoch. Stage 1 is trained for $20$ epochs, and all its parameters are frozen during stage 2. Stage 2 is then trained for $20$ epochs with $T_0$ predicted by the pretrained Stage~1. For stage 1, the 2D backbone is implemented as a 4-stage ResNet~\cite{resnet} with FPN~\cite{fpn}, with output dimensions $\{128,128,256,512\}$. The point cloud backbone is implemented as a 3-stage KPFCNN~\cite{thomas2019kpconv}, with dimensions $\{128,256,512\}$, yielding $\hat{C}=512$ at the coarsest level. For stage 2, the point cloud features used for Gaussian feature splatting are projected to a $128$-dimensional feature space, \textit{i.e.}, ${C}_{Gs}=128$. The Pixel-Aligned Interaction Transformer module is implemented with shifted-window bidirectional cross-attention using a window size of $32$. The image and point cloud fusion heads $\phi_{\mathrm I}$ and $\phi_{\mathrm P}$ are implemented as separate FFNs, each composed of a normalization layer, two linear projections, and a \textit{GELU} activation. The input image resolution is $480\times640$, and the coarsest feature grid has resolution $34\times45$.
%%%%%%%%%%%%%%%%%%%%%%%%%%%%%%%%%%%
%
% -----------------
\section{Experimental Validation}\label{sec_expe}
% -----------------
%
To evaluate the performance of \emph{GRIP}, we follow state of the art image to point cloud registration methods~\cite{wu2024diff,mu2025diff2i2p,cheng2025bridge} and report results on the RGB-D Scenes V2~\cite{rgbdv2} and 7-Scenes~\cite{7scenes} datasets under the Registration Recall (RR), Inlier Ratio (IR), and Feature Matching Recall (FMR) metrics.  % defined in Sec.~\ref{sec_metrics}.
%
% ------------
\subsection{Datasets}
\noindent\textbf{RGB-D Scenes V2.}
Introduced by Lai et al.~\cite{rgbdv2}, RGB-D Scenes V2 contains 14 indoor RGB-D video sequences, covering both tabletop objects and large furniture. Following prior works~\cite{li20232d3d,wu2024diff,mu2025diff2i2p}, we construct the 2D-3D registration benchmark by fusing one point cloud fragment from every 25 consecutive depth frames and pairing it with one RGB image sampled every 25 frames. Only pairs with an overlap ratio of at least $30\%$ are retained. Scenes 1-8 are used for training, scenes 9 and 10 for validation, and scenes 11-14 for testing. This split yields 1,748 training pairs, 236 validation pairs, and 497 test pairs.
\par\noindent\textbf{7-Scenes.}
Similarly, Glocker et al.~\cite{7scenes} introduced a challenging indoor RGB-D dataset with greater scale and viewpoint variations across scenes than RGB-D Scenes V2. The dataset consists of 46 tracked RGB-D sequences captured from 7 indoor scenes using a handheld Kinect RGB-D camera at a resolution of $640 \times 480$. We follow the same procedure as above to process the input images and point clouds. Following~\cite{li20232d3d,mu2025diff2i2p}, we retain only image-point cloud pairs with at least $50\%$ overlap. Using the official sequence split yields 4048 training pairs, 1011 validation pairs, and 2304 testing pairs.
\subsection{Results}
\subsubsection{RGB-D Scenes V2} As reported in Table~\ref{table_rgbdv2_mean}, \emph{GRIP} achieves the best matching performance on RGB-D Scenes V2 in terms of inlier ratio (IR). In particular, \emph{GRIP} reaches an IR of 60.9\%, outperforming the second-best method, R$^{23}$Net~\cite{cheng2026rethinking}, by 17.5 percentage points. Compared with~\cite{li20232d3d}, which obtains 32.4\% IR, this corresponds to a relative improvement of approximately 88\%, demonstrating the effectiveness of the proposed rendering-based refinement for producing geometrically consistent correspondences.

Regarding registration, following prior work~\cite {li20232d3d,wu2024diff}, we report RR at a 10 cm threshold as the primary metric. \emph{GRIP} ranks second overall on this metric, reaching 84.6\% RR with EPro-PnP and 84.9\% RR with score-guided RANSAC, while Diff-Reg~\cite{wu2024diff}, which adopts a diffusion process, obtains 85.7\%. This indicates that the substantial gains in correspondence quality translate into highly competitive registration recall. Moreover, although \emph{GRIP} has a slightly higher network forward time than Diff-Reg, its total inference time is lower because the final pose estimation stage is substantially more efficient. Under identical hardware conditions, \emph{GRIP} reduces total inference latency by 40.2\% with EPro-PnP and by 27.8\% with score-guided RANSAC compared to Diff-Reg (Table~\ref{tab_time}).
\subsubsection{7-Scenes.} 
\begin{figure}[!h]
\centerline{\includegraphics[width=0.8\columnwidth]{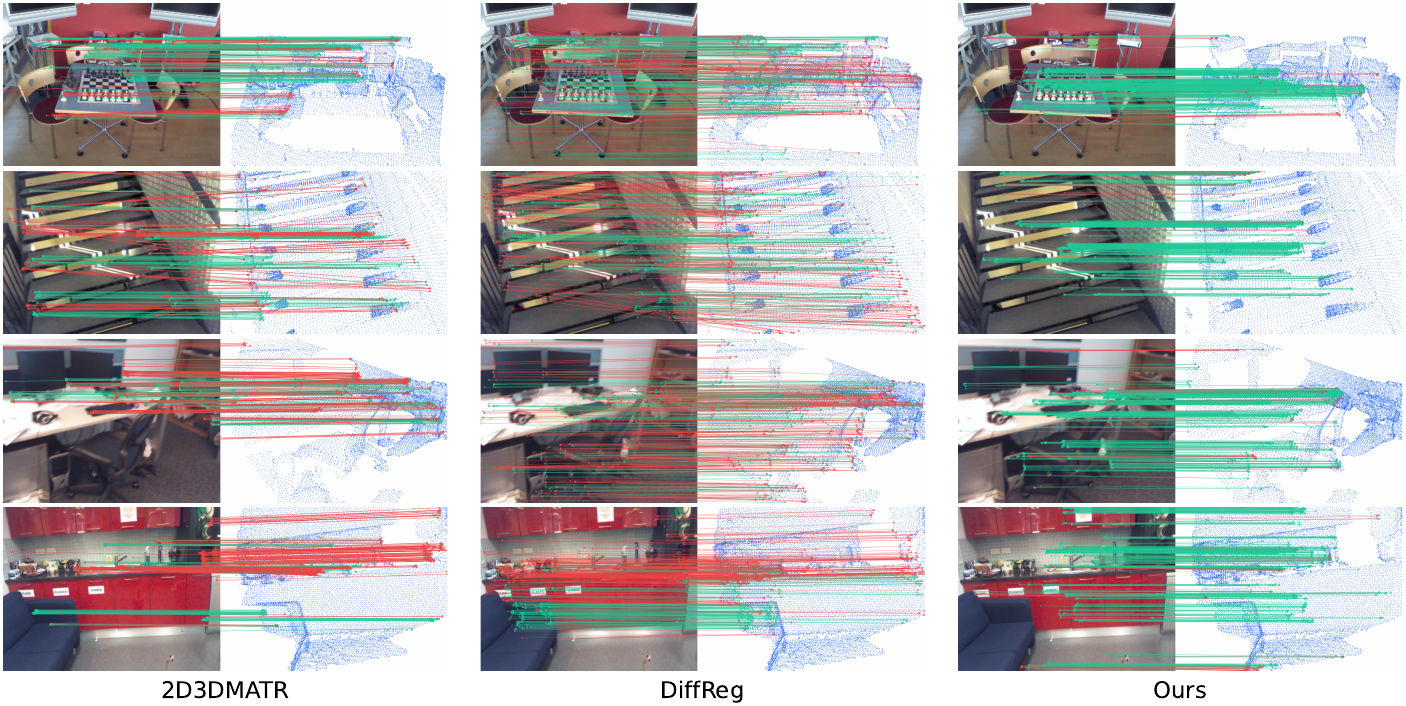}}
\caption{Qualitative comparison of matching results on selected 7Scenes examples. Each row shows the same image/point-cloud pair $512$ correspondences with the highest score are represented. Green indicates inlier matches, while red indicates outliers. }
\label{fig_7scenes}
\end{figure}

Similarly, following previous works~\cite{li20232d3d,wu2024diff}, we evaluate the same metrics on the 7-Scenes dataset. The results reported in Table~\ref{table_7scenes_mean} show trends consistent with those observed on RGBD Scenes V2. In particular, \textit{GRIP} achieves the best IR, reaching $71.9\%$. Compared with the 2D3DMATR baseline, which achieves an IR of $50.1\%$, this represents an absolute improvement of $21.8$ percentage points and a relative improvement of approximately $43.5\%$. The second-best method is R$^{23}$Net, at 54.9\%, which is 17.0 percentage points lower than our method. This substantial gain in IR indicates that \textit{GRIP} produces significantly more geometrically consistent 2D-to-3D correspondences, as qualitatively illustrated in Fig.~\ref{fig_7scenes}.  Although the Feature Matching Recall (FMR) of \emph{GRIP} is slightly lower than the best reported value, its substantially higher IR indicates that the predicted correspondences are more geometrically consistent once a sufficient set of matches is established. This improved correspondence quality directly benefits pose estimation. \textit{GRIP} achieves an RR of $84.1\%$ with EPro-PnP and $84.2\%$ with score-guided RANSAC, outperforming all compared methods. These results show that the proposed method not only increases the number of tentative matches but also improves their geometric accuracy, leading to more robust and successful registration.  

\begin{figure}[t]
    \centering

    \begin{subfigure}{0.48\columnwidth}
        \centering
        \includegraphics[width=0.9\linewidth]{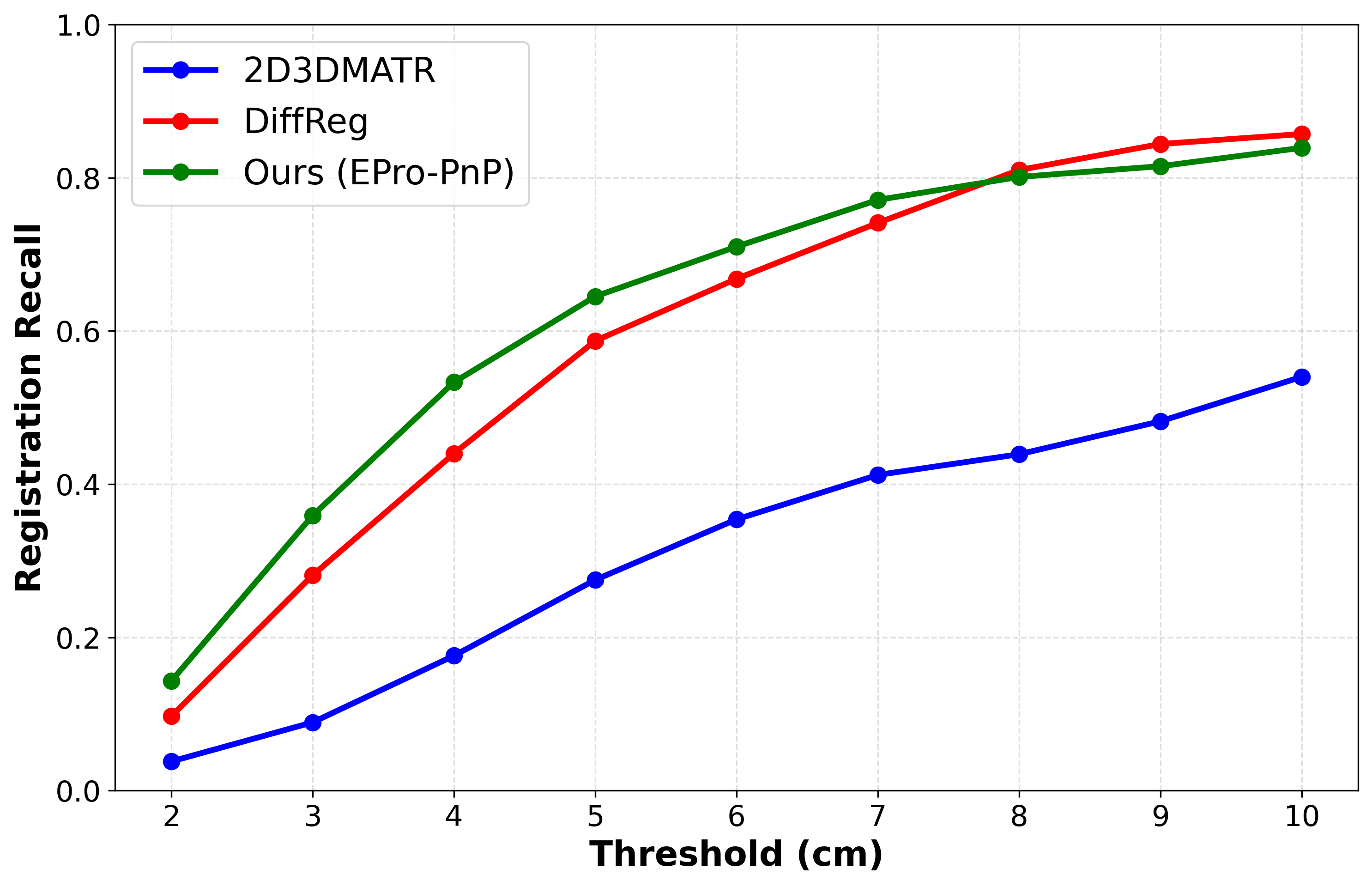}
        \caption{Registration Recall evolution on RGBD V2.}
        \label{fig_courbe_rgbd}
    \end{subfigure}
    \hfill
    \begin{subfigure}{0.48\columnwidth}
        \centering
        \includegraphics[width=0.9\linewidth]{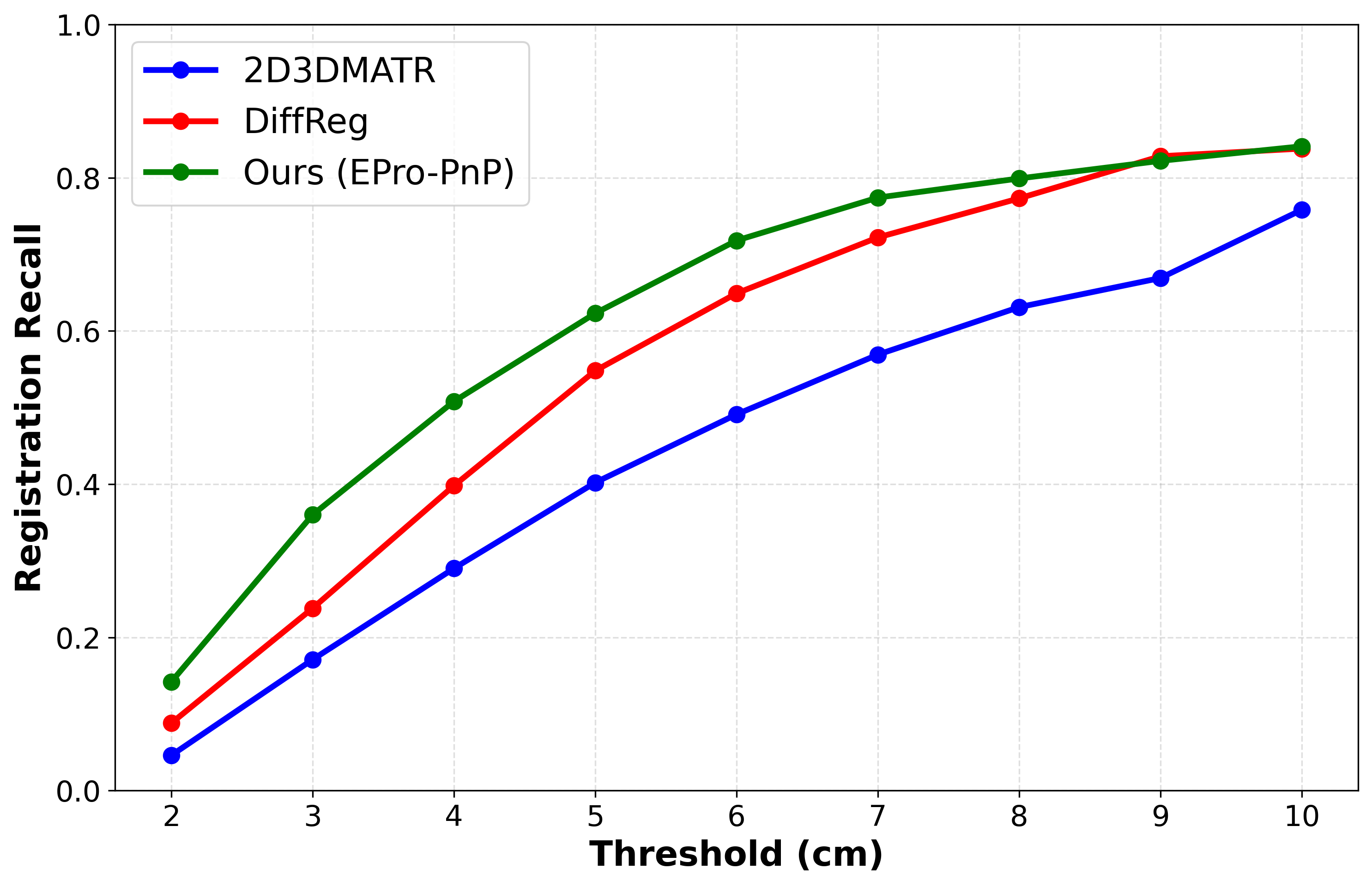}
        \caption{Registration Recall evolution on 7Scenes.}
        \label{fig_courbe_7scenes}
    \end{subfigure}

    \caption{Registration Recall for increasing threshold on RGBD V2 and 7Scenes.}
    \label{fig_courbes_rr}
\end{figure}

In addition, we compute the RR for both datasets using thresholds ranging from 2 cm to 10 cm. As shown in Fig.~\ref{fig_courbes_rr}, \emph{GRIP} achieves higher RR under stricter thresholds,  demonstrating its superior registration precision. Under more relaxed thresholds, Diff-Reg gradually closes the gap with \emph{GRIP} and surpasses it on RGBD Scenes V2, regardless of the pose estimator used.
\begin{table}[t]
\centering
\small
\setlength{\tabcolsep}{3pt}
\renewcommand{\arraystretch}{1.05}

\begin{minipage}{0.48\columnwidth}
\centering
\caption{Evaluation results on RGB-D V2. Bold numbers highlight the best performance; the second best are underlined.}
% \centering}
\resizebox{\linewidth}{!}{%
\begin{tabular}{l|ccc}
\hline

\hline

\hline

Model & IR $\uparrow$ & FMR $\uparrow$ & RR $\uparrow$ \\
\hline
%FCGF-2D3D~\cite{choy2019fully} & 8.1 & 27.1 & 30.4 \\
P2-Net~\cite{wang2021p2} & 12.2 & 59.6 & 38.4 \\
Predator-2D3D~\cite{huang2021predator} & 15.7 & 65.9 & 30.2 \\
2D3D-MATR~\cite{li20232d3d} & 32.4 & 90.8 & 56.4 \\
Diff$^2$I2P~\cite{mu2025diff2i2p} & 36.9 & 77.1 & 60.5 \\
Diff-Reg~\cite{wu2024diff} & 37.7 & 91.4 & \textbf{85.7} \\
$R^{23}$Net~\cite{cheng2026rethinking} & \underline{43.4} & \textbf{93.6} & 77.0 \\
\hline
\textbf{\emph{GRIP} - EPro-PnP} & \textbf{60.9} & \underline{91.8} & \underline{84.6} \\
\textbf{\emph{GRIP} - RANSAC*} & \textbf{60.9} & \underline{91.8} & \underline{84.9} \\
\hline

\hline

\hline

\end{tabular}%
}
\label{table_rgbdv2_mean}
\end{minipage}
\hfill
\begin{minipage}{0.48\columnwidth}
\centering
\caption{Evaluation results on 7-Scenes. Bold numbers highlight the best performance; the second best are underlined.}
\resizebox{\linewidth}{!}{%
\begin{tabular}{l|ccc}
\hline

\hline

\hline

Model & IR $\uparrow$ & FMR $\uparrow$ & RR $\uparrow$ \\
\hline
%FCGF-2D3D~\cite{choy2019fully} & 22.8 & 78.8 & 61.4 \\
P2-Net~\cite{wang2021p2} & 31.7 & 79.0 & 65.7 \\
Predator-2D3D~\cite{huang2021predator} & 23.4 & 77.5 & 48.5 \\
2D3D-MATR~\cite{li20232d3d} & 50.1 & 92.1 & 75.8 \\
Diff$^2$I2P~\cite{mu2025diff2i2p} & 53.2 & \underline{92.2} & 83.0 \\
Diff-Reg~\cite{wu2024diff} & 52.0 & 91.6 & \underline{83.8} \\
$R^{23}$Net~\cite{cheng2026rethinking} & \underline{54.9} & \textbf{93.2} &  \underline{83.8}  \\
\hline
\textbf{\emph{GRIP} - EPro-PnP} & \textbf{71.9} & 89.8 & \textbf{84.1} \\
\textbf{\emph{GRIP} - RANSAC*} & \textbf{71.9} & 89.8 & \textbf{84.2} \\
\hline

\hline

\hline

\end{tabular}%
}
\label{table_7scenes_mean}
\end{minipage}

\vspace{1mm}
\end{table}

\begin{table}[!h]
    \scriptsize
    \caption{Inference time cost comparison on RGBD V2 dataset.}
    \label{tab_time}
    \centering
    % \resizebox{\columnwidth}{!}{%
    \begin{tabular}{c|c|cc|c}
        \hline
        
        \hline
        
        \hline

        \bfseries Method & Registration Recall & \multicolumn{3}{c}{ Mean inference time (seconds)} \\
        \cline{3-5}
        & &  \bfseries Model  & \bfseries Pose  & \bfseries Total   \\
        \hline
        
        Diff-Reg~\cite{wu2024diff} & \textbf{85.7} &  0.702  & 0.634  &  1.336 \\
        \hline   
        \textbf{\emph{GRIP} - EPro-PnP (Ours)}&  84.6 & 0.790  &  0.009 &  \textbf{0.799} \\
        \textbf{\emph{GRIP} - RANSAC*(Ours)} &  84.9 & 0.790  &  0.175 &  \textbf{0.965} \\

        \hline
        
        \hline
        
        \hline
    \end{tabular}
    % }
\end{table}

%%%%%%%%%%%%%%%%%%%%%%%%
\subsection{Ablation Study}

Table~\ref{tab_ablation} shows that the complete GRIP pipeline improves IR/RR from $44.2/78.1$ with the DINOv2 based Stage~1 to $60.9/84.9$, corresponding to gains of $+16.7$ IR and $+6.8$ RR. The ablations indicate complementary contributions from Gaussian splatting, PAIT, and hierarchical decoding. PAIT provides the largest improvement by enabling cross modal interaction on the rendered feature grid, while the hierarchical decoder propagates the refined representation to fine matching. Replacing Gaussian splatting with hard point to pixel projection slightly decreases IR/RR to $59.3/82.3$, with no substantial runtime reduction ($882$\,ms vs.\ $895$\,ms). Overall, these results support the contribution of each component to the complete refinement pipeline.

\begin{table}[!t]
\scriptsize
\setlength{\tabcolsep}{3pt}
\renewcommand{\arraystretch}{1.05}
\caption{Ablation on RGB-D V2.}\label{tab_ablation}

\begin{tabular}{lccccc|cc}
\toprule
Variant & DINO & Feat. Spl. & PAIT & Hier. Dec. & Solver & IR & RR \\
\midrule
Baseline
& \cellcolor{lightred}\xmark & \cellcolor{lightred}\xmark & \cellcolor{lightred}\xmark & \cellcolor{lightred}\xmark &\cellcolor{lightred}RANSAC
& 32.4 & 56.4 \\

Our Stage~1
& \cellcolor{lightgreen}\cmark & \cellcolor{lightred}\xmark & \cellcolor{lightred}\xmark & \cellcolor{lightred}\xmark &\cellcolor{lightred}RANSAC
& 44.2 & 78.1 \\

Our Stage~1
& \cellcolor{lightgreen}\cmark & \cellcolor{lightred}\xmark & \cellcolor{lightred}\xmark & \cellcolor{lightred}\xmark &\cellcolor{lightgreen}Score-guided
& 44.2 & 77.3 \\

Stage~2 w/o PAIT
& \cellcolor{lightgreen}\cmark & \cellcolor{lightgreen}\cmark & \cellcolor{lightred}\xmark & \cellcolor{lightgreen}\cmark &\cellcolor{lightred}RANSAC
& 42.5 & 64.7 \\

GRIP RANSAC
& \cellcolor{lightgreen}\cmark & \cellcolor{lightgreen}\cmark & \cellcolor{lightgreen}\cmark & \cellcolor{lightgreen}\cmark &\cellcolor{lightred}RANSAC
& \textbf{60.9} & 83.9 \\

GRIP wo dec.
& \cellcolor{lightgreen}\cmark & \cellcolor{lightgreen}\cmark & \cellcolor{lightgreen}\cmark & \cellcolor{lightred}\xmark &\cellcolor{lightgreen}Score-guided
& 52.4 & 83.9 \\

GRIP + Hard 
& \cellcolor{lightgreen}\cmark & \cellcolor{lightred}Hard & \cellcolor{lightgreen}\cmark & \cellcolor{lightgreen}\cmark &\cellcolor{lightgreen}Score-guided
& 59.3 & 82.3 \\

GRIP 
& \cellcolor{lightgreen}\cmark & \cellcolor{lightgreen}\cmark & \cellcolor{lightgreen}\cmark & \cellcolor{lightgreen}\cmark &\cellcolor{lightgreen}Score-guided
& \textbf{60.9} & \textbf{84.9} \\

\midrule
Oracle (1-stage)
& \cellcolor{lightgreen}\cmark & \cellcolor{lightgreen}\cmark & \cellcolor{lightgreen}\cmark & \cellcolor{lightgreen}\cmark &\cellcolor{lightred}RANSAC
& 99.0 & 100.0 \\

Oracle (2-stage)
& \cellcolor{lightgreen}\cmark & \cellcolor{lightgreen}\cmark & \cellcolor{lightgreen}\cmark & \cellcolor{lightgreen}\cmark & \cellcolor{lightgreen}Score-guided
& 78.0 & 99.9 \\
\bottomrule
\end{tabular}
\end{table}

\subsection{Discussion and Future Directions}

Experiments on RGB-D Scenes V2 and 7-Scenes demonstrate the effectiveness of \emph{GRIP} for cross-modal image-to-point cloud matching, with particularly strong performance under stricter registration thresholds. For a fair comparison, the results of 2D3D-MATR and Diff-Reg were reproduced using their official implementations in the same experimental environment as \emph{GRIP}. Overall, the results confirm that the proposed rendering-based refinement improves the geometric consistency of the predicted correspondences.

The main limitation of \emph{GRIP} is its dependence on the initial pose. Since feature splatting uses $\mathbf{T}_0$ to project 3D features into the image plane, this initialization must be sufficiently accurate to produce meaningful rendered features. When the initial pose is degenerate or highly inaccurate, the rendered representation becomes poorly aligned with the image observations. As a result, the pixel-aligned transformer may receive uninformative or misleading evidence, which can lead to incorrect correspondences and degrade the final pose. This behavior is particularly evident on the Stairs scene of the 7-Scenes dataset, where the weak baseline initialization performance, i.e., a RR of 28.4\% and an IR of 18.1\%, also limits the improvement achieved by \emph{GRIP}, which reaches a RR of 31.1\% and an IR of 28.6\%. These results suggest that the proposed method should be viewed as a pose-conditioned rendering-based refinement framework whose performance depends on the quality of the initial pose, rather than as a solution to arbitrary initialization.

A promising direction for future work is therefore to reduce the dependence on a single initial transformation. One possibility is to integrate \emph{GRIP} into a generative or diffusion-based registration framework~\cite{ho2020denoising, wu2024diff, mu2025diff2i2p}, where the pose is progressively refined from a noisy transformation on the $\mathrm{SE(3)}$ manifold. Another direction is to sample multiple pose candidates around the initial transformation. Each candidate can produce its own feature-splatting representation, enabling the model to assess multiple alignment hypotheses simultaneously. This raises the likelihood of obtaining at least one informative rendering, effectively broadening the convergence basin and mitigating the impact of poor initializations.

% A promising direction is to reduce the dependence on a single initial transformation, for example through generative pose refinement or multiple pose hypotheses. Such strategies could provide several candidate renderings and improve robustness to inaccurate initialization.

\subsubsection*{Acknowledgments}
This work was supported by the French ANR program MARSurg (ANR-21-CE19-0026).\\
This work was performed using HPC resources from GENCI-IDRIS (Grant 2026-AD011015228R1)

%\TODO{architecture} ==> pas de place :-(
\section{Conclusion}

In this paper, we introduced \emph{GRIP}, a pose-conditioned rendering-based refinement framework for 2D-3D feature matching and registration. The main objective of our approach is to reduce the structural gap between grid-based image descriptors and unordered point cloud descriptors. To this end, we proposed a feature-splatting module that renders learned 3D point features into the image plane, producing an image-aligned point-derived feature map. A pixel-aligned interaction transformer then jointly refines image features and 3D-derived rendered features via bidirectional cross-modal attention on a shared 2D grid. The refined rendered features are finally sampled back to the 3D nodes, enabling improved dense 2D-3D correspondence estimation and final pose refinement. The refined representation improves dense correspondence quality, achieving the highest IR on both evaluated benchmarks and competitive RR, particularly under stricter thresholds. Future work will focus on reducing the dependence on the initial pose through multi-hypothesis or generative refinement.

% Experiments on image-to-point cloud registration benchmarks show that \emph{GRIP} improves correspondence reliability, achieving the highest inlier ratio on both evaluated datasets. These stronger correspondences lead to competitive registration recall on RGB-D Scenes V2 and the best recall on 7-Scenes, confirming the benefit of performing cross-modal interaction in a pixel-aligned feature space. Future work will focus on reducing the dependence on the initial pose, for example, through multi-hypothesis rendering or generative refinement on $\mathrm{SE(3)}$. The introduced principle could also be extended to RGB-guided point cloud registration when aligned visual observations are available.

%
\bibliographystyle{splncs04}
\bibliography{main}
\end{document}